\documentclass[journal]{IEEEtran} 
\usepackage{subfig}
\usepackage[english]{babel}
\usepackage{amsmath}
\usepackage{graphicx}
\usepackage[round]{natbib}
\usepackage{soul}
\usepackage{bm}
\usepackage{balance}
\usepackage{color}
\usepackage[hyphens]{url}

\begin{document}

\title{Born to Learn: the Inspiration, Progress, and Future of Evolved Plastic Artificial Neural Networks}

\author{Andrea Soltoggio\thanks{Department of Computer Science, Loughborough University, LE11 3TU, Loughborough, UK, a.soltoggio@lboro.ac.uk}, Kenneth O. Stanley\thanks{Department of Computer Science, University of Central Florida, Orlando, FL, USA, kstanley@cs.ucf.edu}, Sebastian Risi\thanks{IT University of Copenhagen, Copenhagen, Denmark, sebr@itu.dk}} 

\maketitle 
\begin{abstract}
Biological neural networks are systems of extraordinary computational capabilities shaped by evolution, development, and lifelong learning. The interplay of these elements leads to the emergence of biological intelligence. Inspired by such intricate natural phenomena, Evolved Plastic Artificial Neural Networks (EPANNs) employ simulated evolution in-silico to breed plastic neural networks with the aim to autonomously design and create learning systems. EPANN experiments evolve networks that include both innate properties and the ability to change and learn in response to experiences in different environments and problem domains. EPANNs' aims include autonomously creating learning systems, bootstrapping learning from scratch, recovering performance in unseen conditions, testing the computational advantages of particular neural components, and deriving hypotheses on the emergence of biological learning. Thus, EPANNs may include a large variety of different neuron types and dynamics, network architectures, plasticity rules, and other factors. While EPANNs have seen considerable progress over the last two decades, current scientific and technological advances in artificial neural networks are setting the conditions for radically new approaches and results. Exploiting the increased availability of computational resources and of simulation environments, the often challenging task of hand-designing learning neural networks could be replaced by more autonomous and creative processes. This paper brings together a variety of inspiring ideas that define the field of EPANNs. The main methods and results are reviewed. Finally, new opportunities and possible developments are presented.  
\end{abstract}



\vspace{-4pt}
\section{Introduction}

Over the course of millions of years, evolution has led to the emergence of innumerable biological systems, and intelligence itself, crowned by the evolution of the human brain. Evolution, development, and learning are the fundamental processes that underpin biological intelligence. Thus, it is no surprise that scientists have tried to engineer artificial systems to reproduce such phenomena \citep{sanchez1996phylogeny,sipperSanchezMangeTomassiniPerez-UribeStauffer1997,dawkins2003evolution}. The fields of artificial intelligence (AI) and artificial life (AL) \citep{langton1997artificial} are inspired by nature and biology in their attempt to create intelligence and forms of life from human-designed computation: the main idea is to abstract the principles from the medium, i.e., biology, and utilize such principles to devise algorithms and devices that reproduce properties of their biological counterparts.

One possible way to design complex and intelligent systems, compatible with our natural and evolutionary history, is to simulate natural evolution in-silico, as in the field of evolutionary computation \citep{holland1975adaptation,eiben2015evolutionary}. Sub-fields of evolutionary computation such as evolutionary robotics \citep{harvey1997evolutionary,nolfi2000evolutionary},  learning classifier systems \citep{lanzi2003learning,butz2015learning}, and neuroevolution \citep{yao1999} specifically research algorithms that, by exploiting artificial evolution of physical, computational, and neural models, seek to discover principles behind intelligent and learning systems. 

In the past, research in evolutionary computation, particularly in the area of neuroevolution, was predominantly focused on the evolution of static systems or networks with fixed neural weights: evolution was seen as an alternative to learning rules to search for optimal weights in an artificial neural network (ANN). Also, in traditional and deep ANNs, learning is often performed during an initial training phase, so that weights are static when the network is deployed. Recently, however, inspiration has originated more strongly from the fact that intelligence in biological organisms considerably relies on powerful and general learning algorithms, designed by evolution, that are executed during both development and continuously throughout life. 

As a consequence, the field of neuroevolution is now progressively moving towards the design and evolution of lifelong learning plastic neural systems, capable of discovering learning principles during evolution, and thereby able to acquire knowledge and skills through the interaction with the environment  \citep{coleman2012evolving}. This paper reviews and organizes the field that studies \emph{evolved plastic artificial neural networks}, and introduces the acronym EPANN. EPANNs are \emph{evolved} because parts of their design are determined by an evolutionary algorithm; they are \emph{plastic} because parts of their structures or functions, e.g.\ the connectivity among neurons, change at various time scales while experiencing sensory-motor information streams. The final capabilities of such networks are autonomously determined by the combination of evolved genetic instructions and learning that takes place as the network interacts with an environment. 

EPANNs' ambitious motivations and aims, centered on the autonomous discovery and design of learning systems, also entail a number of research problems. One problem is how to set up evolutionary experiments that can discover learning, and then to understand the subsequent interaction of dynamics across the evolutionary and learning dimensions. A second open question concerns the appropriate neural model abstractions that may capture essential computational principles to enable learning and, more generally, intelligence. One further problem is the size of very large search spaces, and the high computational cost required to simulate even simple models of lifelong learning and evolution. Finally, experiments to autonomously discover intelligent learning systems have a wide range of performance metrics, as their objectives are sometimes loosely defined as the increase of behavioral complexity, intelligence, adaptability, evolvability \citep{miconi2008road}, and general learning capabilities \citep{tonelli2011using}. Thus, EPANNs explore a  larger search space, and address broader research questions, than machine learning algorithms specifically designed to improve performance on well-defined and narrow problems.

The power of EPANNs, however, derives from two autonomous search processes: evolution and learning, which arguably place them among the most advanced AI and machine learning systems in terms of open-endedness, autonomy, potential for discovery, creativity, and human-free design. These systems rely the least on pre-programmed instructions because they are designed to autonomously evolve while interacting with a real or simulated world. Plastic networks, in particular recurrent plastic networks, are known for their computational power \citep{cabessa2014super}: evolution can be a valuable tool to explore the power of those computational structures. 

In recent years, progress in a number of relevant areas has set the stage for renewed advancements of EPANNs: ANNs, in particular deep networks, are becoming increasingly more successful and popular; there has been a remarkable increase in available computational power by means of parallel GPU computing and dedicated hardware; a better understanding of search, complexity, and evolutionary computation allows for less naive approaches; and finally, neuroscience and genetics provide us with an increasingly large set of inspirational principles. This progress has changed the theoretical and technological landscape in which EPANNs first emerged, providing greater research opportunities than in the past.

Despite a considerable body of work, research in EPANNs has never been unified through a single description of its motivations and inspiration, achievements and ambitions. This paper aims firstly to outline the inspirational principles that motivate EPANNs (Section \ref{sec:inspiration}). The main properties and aims of EPANNs, and suitable evolutionary algorithms are presented in Section \ref{sec:3}. The body of research that advanced EPANNs is brought together and described in Section \ref{sec:EAPB}. Finally, the paper outlines new research directions, opportunities, and challenges for EPANNs (Section \ref{sec:OC}).

\section{Inspiration}
\label{sec:inspiration}

EPANNs are inspired by a particularly large variety of ideas from biology, computer science, and other areas \citep{floreano2008bio,downing2015intelligence}. It is also the nature of inspiration to be subjective, and some of the topics described in this section will resonate differently to different readers. We will touch upon large themes and research areas with the intent to provide the background and motivations to introduce the properties, the progress, and the future directions of EPANNs in the remainder of the paper. 

The precise genetic make-up of an organism, acquired through millions of years of evolution, is now known to determine the ultimate capabilities of complex biological neural systems \citep{deary2009genetic,hopkins2014chimpanzee}: different animal species manifest different levels of skills and intelligence because of their different genetic blueprint \citep{schreiweis2014humanized}. The intricate structure of the brain emerges from one single zygote cell through a developmental process \citep{kolb2011brain}, which is also strongly affected by input-output learning experiences throughout early life \citep{hensch1998local,kolb2011brain}. Yet high levels of plasticity are maintained throughout the entire lifespan \citep{merzenich1984somatosensory,kiyota2017neurogenesis}. These dimensions, \emph{evolution}, \emph{development} and \emph{learning}, also known as the phylogenetic (evolution), ontogenetic (development) and epigenetic (learning) (POE) dimensions \citep{sipperSanchezMangeTomassiniPerez-UribeStauffer1997}, are essential for the emergence of biological plastic brains.

The POE dimensions lead to a number of research questions. Can artificial intelligence systems be entirely engineered by humans, or do they need to undergo a less human-controlled process such as evolution? Do intelligent systems need to learn, or could they be born already knowing? Is there an optimal balance between innate and acquired knowledge? Opinions and approaches are diverse.  Additionally, artificial systems do not need to implement the same constraints and limitations as biological systems \citep{bullinaria2003biological}. Thus, inspiration is not simple imitation. 

{EPANNs assume that both evolution and learning, if not strictly necessary, are conducive to the emergence of a strongly bio-inspired artificial intelligence.} While artificial evolution is justified by the remarkable achievements of natural evolution, the role of learning has gathered significance in recent years. We are now more aware of the high level of brain plasticity, and its impact on the manifestation of behaviors and skills \citep{ledoux2003synaptic,doidge2007brain,grossberg2012studies}. Concurrently, recent developments in machine learning \citep{michalski2013machine,alpaydin2014introduction} and neural learning \citep{deng2013new,lecun2015deep, silver2016mastering}, have  highlighted the importance of learning from large input-output data and extensive training. Other areas of cognition such as the capabilities to make predictions \citep{hawkins2007intelligence}, to establish associations \citep{rescorla2014pavlovian} and to regulate behaviors \citep{carver2012attention} are also based on learning from experience. Interestingly, skills such as reading, playing a musical instrument, or driving a car, are mastered even if none of those behaviors existed during evolutionary time, and yet they are mostly unique to humans. Thus, human genetic instructions have evolved not to learn specific tasks, but to synthesize recipes to learn a large variety of general skills. We can conclude that the evolutionary search of learning mechanisms in EPANNs tackles both the long-running \emph{nature vs.\ nurture} debate \citep{moore2003dependent}, and the fundamental AI research that studies learning algorithms. This review focuses on \emph{evolution} and \emph{learning}, and less on development, which can be interpreted as a form of learning if affected by sensory-motor signals. We refer to \cite{stanley2003taxonomy} for an overview of artificial developmental theories.

\begin{figure*}[t]
\begin{centering}
\includegraphics[width=0.9\textwidth]{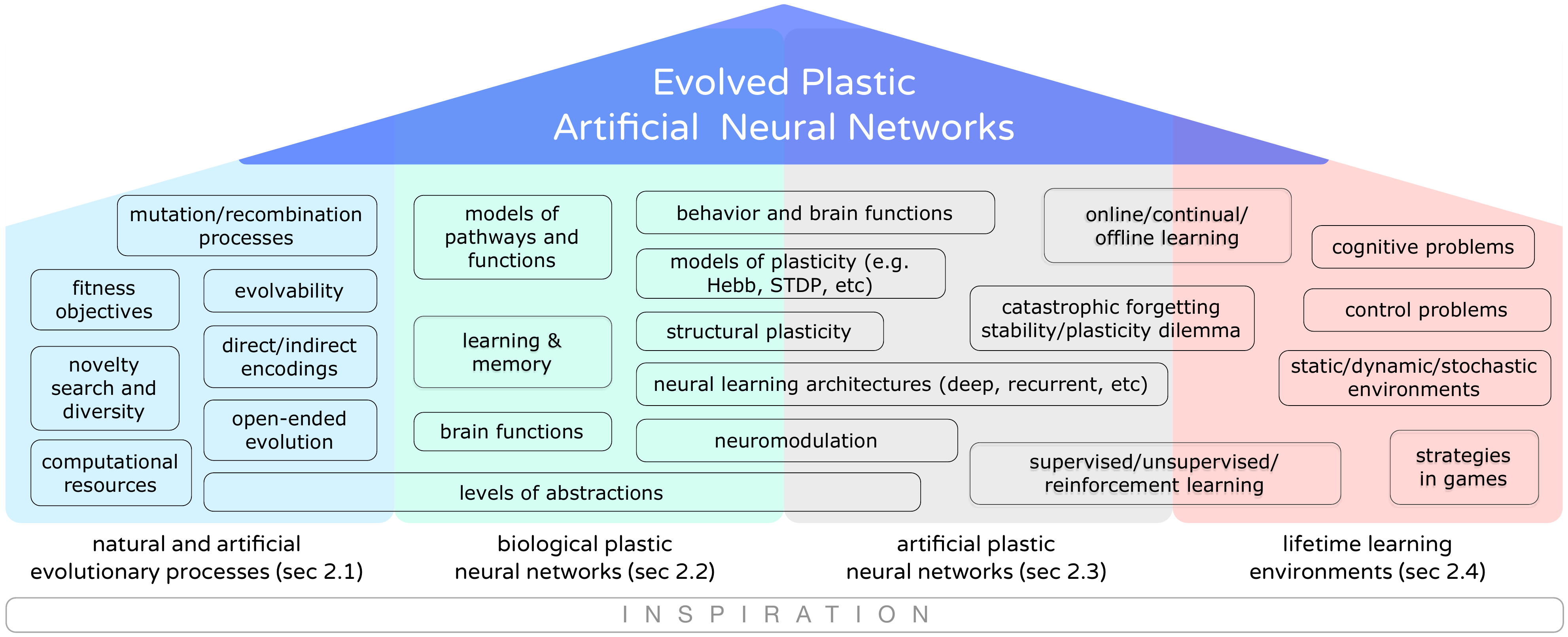}
\caption{EPANN inspiration principles described in Section 2. The figure brings together the variety of inspirational topics and areas with no pretense of taxonomic completeness.}
\label{fig.B2Linspiration}
\end{centering}
\end{figure*}

Whilst the range of inspiring ideas is large and heterogeneous, the analysis in this review proposes that such ideas can be grouped under the following areas:

\begin{itemize}
\item \emph{natural and artificial evolutionary processes},
\item \emph{plasticity in biological neural networks},
\item \emph{plasticity in artificial neural networks}, and 
\item \emph{natural and artificial learning environments}.
\end{itemize}
Figure \ref{fig.B2Linspiration} graphically summarizes the topics described in sections \ref{sec:natural_and_artificial_evo}-\ref{sec:lifelong} from which EPANNs take inspiration. 

\subsection{Natural and artificial evolutionary processes}
\label{sec:natural_and_artificial_evo}

A central idea in evolutionary computation \citep{goldberg1988genetic} is that evolutionary processes, similar to those that occurred in nature during the course of billions of years \citep{darwin1859,dobzhansky1970genetics}, can be simulated with computer software. This idea led to the belief that intelligent computer programs could emerge with little human intervention by means of evolution in-silico   \citep{holland1977cognitive,koza1992genetic,fogel2006evolutionary}.

The emergence of evolved intelligent software, however, did not occur as easily as initially hoped. The reasons for the slow progress are not completely understood, but a number of problems have been identified, likely related to the simplicity of the early implementations of evolutionary algorithms and the high computational requirements. Current topics of investigation focus on levels of abstraction, diversity in the population, selection criteria, the concepts of evolvability and scalability \citep{wagner1996perspective,pigliucci2008evolvability,lehman2013evolvability}, the encoding of genetic information through the genotype-phenotype mapping processes \citep{wagner1996perspective,hornby2002creating}, the deception of fitness objectives, and how to avoid them \citep{lehman2008exploiting,stanley2015greatness}. It is also not clear yet which stepping stones were most challenging for natural evolution \citep{roff1993evolution,stanley2015greatness} in the evolutionary path to intelligent and complex forms of life. This lack of knowledge highlights that our understanding of natural evolutionary processes is incomplete, and thus the potential to exploit computational methods is not fully realized. In particular, EPANN research is concerned with those evolutionary algorithms that allow the most creative, open-ended and scalable design. Effective evolutionary algorithms and their desirable features for EPANNs are detailed later in Section \ref{sec:EA}.

\subsection{Plasticity in biological neural networks}
\label{sec:plasticity_in_biological}

Biological neural networks demonstrate lifelong learning, from simple reflex adaptation to the acquisition of astonishing skills such as social behavior and learning to speak one or more languages. Those skills are acquired through experiencing stimuli and actions and by means of learning mechanisms not yet fully understood \citep{bear2007neuroscience}. The brief overview here outlines that adaptation and learning strongly rely on neural plasticity, understood as ``the ability of neurons to change in form and function in response to alterations in their environment'' \citep{Kaas200110542}.

The fact that experiences guide lifelong learning was extensively documented in the works of \emph{behaviorism} by scientists such as \cite{thorndike1911}, \cite{pavlov1927}, \cite{skinner1938behavior,skinner1953}, and \cite{hull1943} who started to test scientifically how experiences cause a change in behavior, in particular as a result of learning associations and observable behavioral patterns \citep{staddon1983}. This approach means linking behavior to brain mechanisms and dynamics, an idea initially entertained by Freud \citep{kvarphippe1983psychology} and later by other illustrious scientists \citep{hebb1949,kandel2007}. A seminal contribution to link psychology to physiology came from \cite{hebb1949}, whose principle that neurons that \emph{fire together, wire together} is relevant to understanding both low level neural wiring and high level behaviors \citep{doidge2007brain}. Much later, a Hebbian-compatible rule that regulates synaptic changes according to the firing times of the presynaptic and postsynaptic neurons was observed by \cite{markram1997regulation} and named Spike-Timing-Dependent Plasticity (STDP).

The seminal work of \cite{kandelTauc1965}, and following studies \citep{clarkKandel1984}, were the first to demonstrate that changes in the strength of connectivity among neurons{, i.e.\ plasticity,} relates to behavior learning. \cite{waltersByrne1983} showed that, by means of plasticity, a single neuron can perform associative learning such as classical conditioning, a class of learning that is observed in simple neural systems such as that of the Aplysia \citep{carewWaltersKandel1981}. Plasticity driven by local neural stimuli, i.e.\ compatible with the Hebb synapse \citep{hebb1949,brownKairissKeenan1990}, is responsible not only for fine tuning, but also for building a working visual system in the cat's visual cortex \citep{rauscheckerSinger1981}.  

{Biological plastic neural networks are also capable of structural plasticity, which creates new pathways among neurons
\citep{lamprecht2004structural,chklovskii2004cortical,russo2010addicted}: it occurs primarily during development, but there is  evidence that it continues well into adulthood \citep{pascual2005plastic}.  Axon growth, known to be regulated by neurotrophic nerve growth factors \citep{tessier1996molecular}, was also modeled computationally in \cite{roberts2014can}.} Developmental processes and neural plasticity are often indistinguishable \citep{kolb1989brain,pascual2005plastic} because the brain is highly plastic during development. Neuroscientific advances reviewed in \cite{damasio1999feeling,ledoux2003synaptic,pascual2005plastic,doidge2007brain,draganski2008training} outline the importance of structural plasticity in learning motor patterns, associations, and ways of thinking. Both structural and functional plasticity in biology are essential to acquiring long-lasting new skills, and for this reason appears to be an important inspiration for EPANNs.

{Finally, an important mechanism for plasticity and behavior is neuromodulation \citep{marderThirumalai2002,gu2002,baileyGiustettoHuangHawkinsKandel2000}.} Modulatory chemicals such as acetylcholine (ACh), norepinephrine (NE), serotonin (5-HT) and dopamine (DA) appear to regulate a large variety of neural functions, from arousal and behavior \citep{harris-warrickMarder1991,hasselmoSchnell1994,marder1996,katz1995, katzFrost1996}, to pattern generation \citep{katzGettingFrost1994}, to memory consolidation \citep{kupfermann1987,hasselmo1995,marder1996,hasselmo1999}. Learning by reward in monkeys was linked to dopaminergic activity during the 1990s with studies by \cite{schultzApicellaLjungberg1993,schultzDayanMontague1997,schultz1998}. For these reasons, neuromodulation is considered an essential element in cognitive and behavioral processes, and has been the topic of a considerable amount of work in EPANNs (Section \ref{sec:NM}). 

{This compact overview suggests that neural plasticity encompasses an important set of mechanisms, regulated by a rich set of signals and dynamics currently mostly ignored in ANNs. Thus, EPANNs can be used to explore, via evolutionary search, the potential of plasticity and to answer questions such as:} (1) How does a brain-like structure form---driven both by genetic instructions and neural activity---and acquire functions and behaviors? (2) What are the key plasticity mechanisms from biology that can be applied to artificial systems such as EPANNs? (3) Can memories, skills, and behaviors be stored in plastic synaptic connections, in patterns of activities, or in a combination of both? Whilst neuroscience continues to provide inspiration and insight into plasticity in biological brains, EPANNs serve the complementary objective of seeking, implementing, and verifying designs of bio-inspired methods for adaptation, learning, and intelligent behavior.

\subsection{Plasticity in artificial neural networks}
\label{sec:plasticity_in_artificial}

{In EPANN experiments, evolution can be seen as a meta-learning process. Thus, established learning rules for ANNs are often used as \emph{ingredients} that evolution uses to search for good parameter configurations, efficient combinations of rules and network topologies, new functions representing novel learning rules, etc.} EPANN experiments are suited to include the largest possible variety of rules because of (1) the variety of possible tasks in a simulated behavioral experiment and (2) the flexibility of evolution to combine rules with no assumptions about their dynamics. The following gives a snapshot of the extent and scope of various learning algorithms for ANN that can be used as building blocks of EPANNs.

In supervised learning, \emph{backpropagation} is the most popular learning rule used to train both shallow and deep networks \citep{rumelhart1988learning,widrow199030,lecun2015deep} for classification or regression. Unsupervised learning is implemented in neural networks with self-organizing maps (SOM) \citep{kohonen1982self,kohonen1990}, auto-encoders \citep{bourlard1988auto}, restricted Boltzmann machines (RBM) \citep{hinton2006reducing}, Hebbian plasticity \citep{hebb1949,gerstner2002mathematical,cooper2005}, generative adversarial networks \citep{goodfellow2014GAN}, and various combinations of the above. RBM learning is considered related to the free-energy principle, proposed by \cite{friston2009free} as a central principle governing learning in the brain. Hebbian rules, in particular, given their biological plausibility and unsupervised learning, are a particularly important inspirational principle for EPANNs. Variations \citep{willshaw1990optimal} have been proposed to include, e.g., terms to achieve stability \citep{oja1982,bienenstockCooperMunro1982} and various constraints \citep{millerMacKay1994}, or more advanced update dynamics such as dual weights for fast and slow decay \citep{levy1995connectionist,hinton1987using,bullinaria2009evolved,soltoggioHTP2014}. Hebbian rules have been recently proposed to minimize defined cost functions \citep{pehlevan2015hebbian,BahrounHunsickerSoltoggio2017a}, and more advanced systems have used backpropagation as meta-learning to tune Hebbian rules \citep{miconi2018differentiable}.

Neuromodulated plasticity \citep{fellousLinster1998} is often used to implement reward-learning in neural networks.  Such a modulation of signals, or gated learning  \citep{abbott1990}, allows for amplification or reduction of signals and has been implemented in numerous models  \citep{baxterCanvierClarkByrne1999,suriBargasArbib2001,birmingham2001,alexanderSporns2002,doya2002,fujiiSaitoNakahiraIshiguro2002,suri2002,ziemkeThieme2002,spornsAlexander2003,krichmar2008neuromodulatory}. 

Plastic neural models are also used to demonstrate how behavior can emerge from a particular circuitry modeled after biological brains. Computational models of, e.g., the basal ganglia and modulatory systems may propose plasticity mechanisms and aim to demonstrate the computational relations among various nuclei, pathways, and learning processes \citep{krichmar2008neuromodulatory,vitayHamkerFCN2010,schroll2015computational}.

Finally, plasticity rules for spiking neural networks \citep{maass2001pulsed} aim to demonstrate unique learning mechanisms that emerge from spiking dynamics \citep{markram1997regulation,izhikevich2006polychronization,izhikevich2007solving}, as well as model biological synaptic plasticity  \citep{gerstner2002spiking}.

Plasticity in neural networks, when continuously active, was also observed to cause catastrophic forgetting \citep{robins1995}. If learning occurs continuously, new information or skills have the potential to overwrite previously acquired information or skills, a problem also known as plasticity-stability dilemma \citep{abraham2005memory,finnie2012role}.

In conclusion, a large range of plasticity rules for neural networks have been proposed to 
solve different problems. In the majority of cases, a careful matching and engineering of rules, architectures and problems is necessary, requiring considerable design effort. The variety of algorithms also reflects the variety of problems and solutions. One key aspect is that EPANN systems can effectively play with all possible plasticity rules to offer a unique testing tool and assess the effectiveness and suitability of different models, or their combination, in a variety of different scenarios.

\subsection{Lifelong learning environments}
\label{sec:lifelong}

One aspect of EPANNs is that they can continuously improve and adapt both at the evolutionary scale and at the lifetime scale in a virtually unlimited range of problems. Natural environments are an inspiration for EPANNs because organisms have evolved to adapt to, and learn in, a variety of conditions. Fundamental questions are: what makes an environment conducive to the evolution of learning and intelligence? What are the challenges faced by learning organisms in the natural world, and how does biological learning cope with those? How can those challenges be abstracted and ported to a simulated environment for EPANNs? EPANNs employ lifelong learning environments in the attempt to provide answers to such questions.

In the early phases of AI, logic and reasoning were thought to be the essence of intelligence \citep{cervier1993ai}, so symbolic input-output mappings were employed as tests. Soon it became evident that intelligence is not only symbol manipulation, but resides also in subsymbolic problem solving abilities emerging from the interaction of brain, body, and environment \citep{steels1993artificial,sims1994evolving}. More complex simulators of real-life environments and body-environment interaction were developed to better represent the \emph{enactivist} philosophy \citep{varela2017embodied} and cognitive theories on the emergence of cognition \citep{butz2016mind}. Other environments focus on high-level planning and strategies required, e.g., when applying AI to games \citep{allis1994searching,millington2016artificial} or articulated robotic tasks. Planning and decision making with high bandwidth sensory-motor information flow such as those required for humanoid robots or self-driving vehicles are current benchmarks for lifelong learning systems. Finally, environments in which affective dynamics and feelings play a role are recognized as important for human well being \citep{deBottonWired2016,lee2005toward}. Those intelligence-testing environments are effectively the ``worlds'' in which EPANNs may evolve and live in embodied forms, and thus largely shape the EPANN design process.

Such different testing environments have very different features, dynamics, and goals that fall into different machine learning problems. For example, supervised learning can be mapped to a fitness function when a precise target behavior exists and is known. If it is useful to find relationships and regularities in the environment, unsupervised learning, representation learning, or modularity can be evolved  \citep{bullinaria2007understanding}. If the environment provides rewards, the objective may be to search for behavioral policies that lead to collecting rewards: algorithms specifically designed to do so are called reinforcement learning  \citep{SuttonBarto1998}. While reinforcement learning maximizes a reward or fitness, recent advances in evolutionary computation \citep{lehman2011abandoning,stanley2015greatness} suggest that it is not always the \emph{fittest}, but at times it is the \emph{novel} individual or behavior that can exploit environmental niches, thus leading to creative evolutionary processes similar to those observed in nature. Temporal dynamics, i.e.\ when a system requires to behave over time according to complex dynamics, need different computational structures from functions with no temporal dynamics. This case is typical for EPANN experiments that may exhibit a large variety of time scales in complex behavioral tasks. With traditional approaches, all those different cases require careful manual design to solve each problem. In contrast, the evolution in EPANNs can be designed to address most problems by mapping a measure of success to a fitness value, thus searching for solutions in an increasingly large variety of problems and environments. In conclusion, lifelong learning environments of different types can be used with EPANNs to explore innovative and creative solutions with limited human intervention and design. 


\section{Properties, aims, and evolutionary algorithms for EPANNs}
\label{sec:3}

Having introduced the inspirational principles of EPANNs, we now  propose: a list of primary properties that define EPANNs (Section \ref{sec:properties}); the principal aims of EPANN studies (Section \ref{sec:aims}); and a list of desired properties of EAs for EPANNs (Section \ref{sec:EA}). 

\subsection{EPANN properties}
\label{sec:properties}

EPANNs, as formalized in this review, are defined as artificial neural networks with the following properties: 

\textbf{Property 1 - Evolution}: \emph{Parts of an EPANN are determined by an evolutionary algorithm.} Inspired by natural and artificial evolution (Section \ref{sec:natural_and_artificial_evo}), such search dynamics in EPANNs implement a design process. 

\textbf{Property 2 - Plasticity}: \emph{Parts of the functions that process signals within the network change in response to signals propagated through the network, and those signals are at least partially affected by stimuli.} Inspired by biological findings on neural plasticity (Section \ref{sec:plasticity_in_biological}) and empowered by the effectiveness of plasticity in neural models (Section \ref{sec:plasticity_in_artificial}), EPANNs either include such mechanisms or are set up with the conditions to evolve them. 

\textbf{Property 3 - Discovery of learning}: \emph{Property 1 and 2 are implemented to discover, through evolution, learning dynamics within an artificial neural network.} Thus, an EPANN uses both evolution and plasticity in synergy to achieve learning. Such a property can be present in different degrees, from the highest degree in which no learning occurs before evolution and it is therefore discovered from scratch, to the lowest degree in which learning is fine-tuned and optimized, e.g, when evolution is seeded with proven learning structures. Given the very diverse interpretations of \emph{learning} in different domains, we refer to \cite{michalski2013machine} for an overview, or otherwise assume the general machine learning definition by \cite{michalski2013machine}\footnote{A computer program is said to learn from experience E with respect to some class of tasks T and performance measure P, if its performance at tasks in T, as measured by P, improves with experience E.}. 

\textbf{Property 4 - Generality}: \emph{Properties 1 to 3 are independent from the learning problem(s) and from the plasticity mechanism(s) that are implemented or evolved in an EPANN.} Exploiting the flexibility of evolution and learning, (1) EPANNs can evolve to solve problems of different nature, complexity, and time scales (Section \ref{sec:lifelong}); (2) EPANNs are not limited to specific learning dynamics because often it is the aim of the experiment to discover the learning mechanism throughout evolution and interaction with the environment. 

In summary, the EPANNs' properties indicate that within simple assumptions, i.e., using plasticity and evolution, EPANNs are set to investigate the design of learning in creative ways for a large variety of learning problems.

\subsection{Aims}
\label{sec:aims}

Given the above properties, EPANN experiments can be set up to achieve the following aims.

\textbf{Aim 1}: \emph{Autonomously design learning systems}: in an EPANN experiment, it is essential to delegate some design choices of a learning system to the evolutionary process, so that the design is not entirely determined by the human expert and can be automated. The following sub-aims can then be identified.

\textbf{Aim 1.1}: \emph{Bootstrap of learning from scratch}: in an EPANN experiment, it may be desirable to initialize the system with no learning capabilities before evolution takes place, so that the best learning dynamics for a given environment is evolved rather than human-designed.

\textbf{Aim 1.2}: \emph{Optimize performance}: as opposed to \emph{Aim 1.1}, it may be desirable to initialize the system with well know learning capabilities, so that evolution can autonomously optimize the system, e.g., for final performance after learning. 

\textbf{Aim 1.3}: \emph{Recover performance in unseen conditions}: in an EPANN experiment, the desired outcome may be to enable the learning system to autonomously evolve from solving a set of problems to another set without human intervention.

\textbf{Aim 2}: \emph{Test the computational advantages of particular neural components}: the aim of an EPANN experiment might be to test whether particular neural dynamics or components have an evolutionary advantage when implementing particular learning functions. The presence of particular neural component may be fostered by evolutionary selection. 

\textbf{Aim 3}: \emph{Derive hypotheses on the emergence of biological 
learning}: an aim may be to draw similarities or suggest hypotheses on how learning evolved in biological systems, particularly in combination with \emph{Aim 1.1} (bootstrap of learning).

Aim 1 is always present in any EPANN because it derives from the EPANN properties. The other aims may be present in different EPANN studies and can be expanded into more detailed and specific research hypotheses.

\subsection{Evolutionary algorithms for EPANNs}
\label{sec:EA}


In contrast to parameter optimization \citep{back1993overview} in which search spaces are often of fixed dimension and static, EPANNs evolve in dynamic search spaces in which learning further increases the complexity of the evolutionary search and of the problem itself.  Evolutionary algorithms \citep{holland1975adaptation,michalewicz1994gas} for EPANNs often require additional advanced features to cope with the challenges of the evolution of learning and open evolutionary design \citep{bentley1999evolutionary}. The analysis in this review suggests that evolutionary algorithms (EAs) for EPANNs may implement the following desirable properties.

\subsubsection{Variable genotype length and growing complexity}
\label{sec:EA:variable}
for some learning problems, the size and properties of a network that can solve them are not known in advance. Therefore, a desirable property of the EA for EPANNs is that of increasing the length of the genotype, and thus the information contained in it, as evolution may discover increasingly more complex strategies and solutions that may require larger networks (see, e.g., \citet{stanley2002evolving}).

\subsubsection{Indirect genotype to phenotype encoding}
\label{sec:EA:indirect}
in nature, phenotypes are expressions of a more compact representation: the genetic code. Similarly, EAs may represent genetic information in a compact form, which is then mapped to a larger phenotype. Although EPANNs do not require such a property, such an approach promises better scalability to large networks (see, e.g., \citet{risi2010indirectly}).

\subsubsection{Expressing regularities, repetitions, and patterns}
\label{sec:EA:regularities}
indirect encodings are beneficial when they can use one set of instructions in the genotype to generate more parts in the phenotype. This may involve expressing regularities like symmetry (e.g.\ symmetrical neural architectures), repetition (e.g.\ neural modules), repetition with variation (similar neural modules), and patterns, e.g., motifs in the neural architecture (see, e.g., \citep{stanley2007compositional}).

\subsubsection{Effective exploration via mutation and recombination}
\label{sec:EA:exploration}
genetic mutation and sexual reproduction in nature allow for the expression of a variety of phenotypes and for the exploration of new solutions, but seldom lead to highly unfit individuals \citep{ay2007robustness}. Similarly, EAs for EPANNs need to be able to effectively mutate and recombine genomes without destroying the essential properties of the solutions. EAs may use recombination to generate new solutions from two parents: how to effectively recombine genetic information from two EPANNs is still an open question (see, e.g., tracking genes through historical marking in NEAT \citep{stanley2002evolving}).

\subsubsection{Genetic encoding of plasticity rules}
\label{sec:EA:ruleEncoding}
just as neural networks need a genetic encoding to be evolved, so do plasticity rules. EPANN algorithms require the integration of such a rule in the genome. The encoding may be restricted to a simple parameter search,  or evolution may search a larger space of arbitrary and general plasticity rules. Plasticity may also be applied to all or parts of the network, thus effectively implementing the evolution of learning architectures. 

\subsubsection{Diversity, survival criteria, and low selection pressure}
\label{sec:EA:diversity}
the variety of solutions in nature seems to suggest that diversity is a key aspect of natural evolution. EAs for EPANNs are likely to perform better when they can maintain diversity in the population, both at the genotype and phenotype levels. Local selection mechanisms were shown to perform well in EPANN experiments \citep{soltoggioPhDThesis2008}. Niche exploration and behavioral diversity \citep{lehman2011abandoning} could also play a key role for creative design processes. Low selection pressure and survival criteria might be crucial to evolve learning in deceptive environments (see Section \ref{sec:interaction_evo_learning}).

\subsubsection{Open-ended evolution}
\label{sec:EA:open-ended}
evolutionary optimization aims to quickly optimize parameters and reach a fitness plateau. On the contrary,  EAs for EPANNs often seek a more open-ended evolution that can evolve indefinitely more complex solutions given sufficient computational power \citep{taylor2016open,stanleyLehmanSoros2017}.

\subsubsection{Implementations}
many EAs include one or more of these desirable properties \citep{fogel2006evolutionary}. Due to the complexity of neural network design, the field of neuroevolution was the first to explore most of those extensions of standard evolutionary algorithms. Popular algorithms include early work of \citet{angeline1994evolutionary} and \citet{yao1997new} to evolve fixed weights (i.e.\ weights that do not change while the agent interacts with its environment) and the topology of arbitrary neural networks, e.g., recurrent (addressing \ref{sec:EA:variable} and \ref{sec:EA:exploration}). Neuroevolution of Augmenting Topologies (NEAT) \citep{stanley2002evolving} leverages three main aspects: a recombination operator intended to preserve network function (addressing \ref{sec:EA:exploration});  speciation (addressing \ref{sec:EA:diversity}); and evolution from small to larger networks (addressing \ref{sec:EA:variable}). Similarly, EPANN-tailored EAs in \cite{soltoggioPhDThesis2008} employ local selection mechanisms to maintain diversity. Analog Genetic Encoding (AGE) \citep{mattiussi2007analog} is a method for indirect genotype to phenotype mapping that can be used in combination with evolution to design arbitrary network topologies (addressing \ref{sec:EA:variable} and \ref{sec:EA:indirect}) and was used with EPANNs. HyperNEAT \citep{stanley2009hypercube} is an indirect representation method that combines the NEAT algorithm with compositional patterns producing networks (CPPN) \citep{stanley2007compositional} (\ref{sec:EA:variable}, \ref{sec:EA:indirect} and \ref{sec:EA:regularities}). Novelty search \citep{lehman2008exploiting,lehman2011abandoning} was introduced as an alternative to the \emph{survival of the fittest} as a selection mechanism (\ref{sec:EA:diversity}). Initially, the majority of these neuroevolution algorithms were not devised to evolve plastic networks, but rather fixed networks in which the final synaptic weights were encoded in the genome. To operate with EPANNs, these algorithms need to integrate additional genotypical instructions to evolve plasticity rules (\ref{sec:EA:ruleEncoding}).

By adding these EA features (\ref{sec:EA:variable} - \ref{sec:EA:open-ended}) to standard evolutionary algorithms, EPANNs aim to search extremely large search spaces in fundamentally different and more creative ways than traditional heuristic searches of parameters and hyper-parameters.

\begin{figure*}
\begin{centering}
\includegraphics[width=0.7\textwidth]{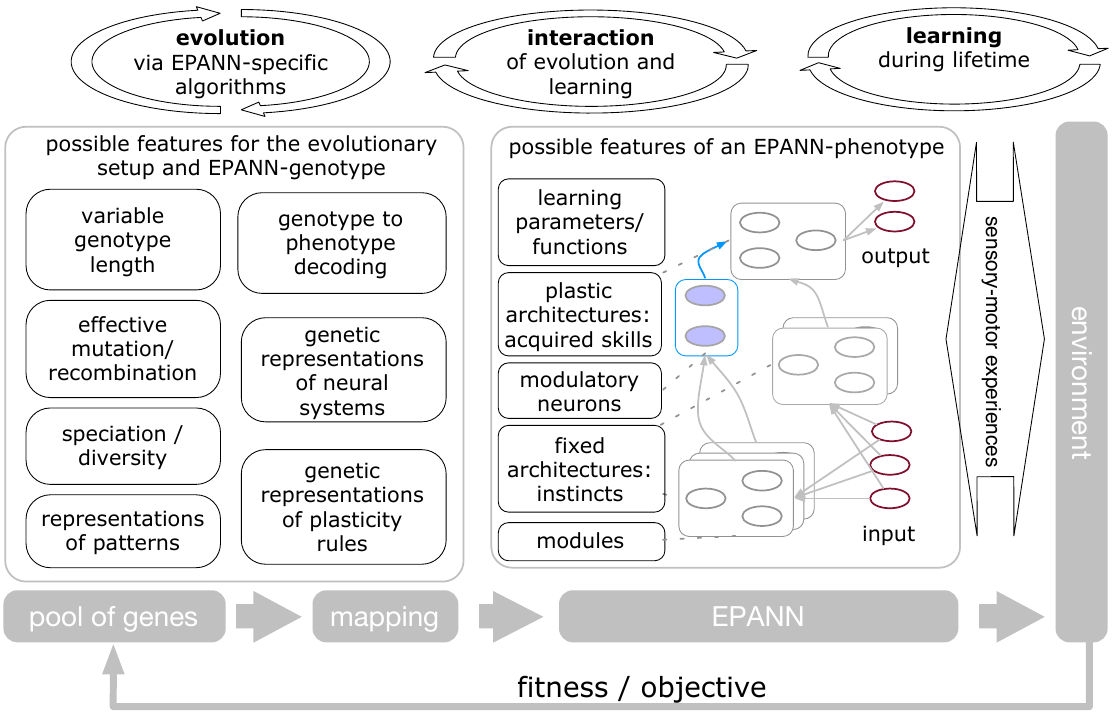}
\caption{Main elements of an EPANN setup in which simulated evolution (left) and an environment (right) allow for an EPANN (center) to evolve through generations and learn within a lifetime in the environment. Possible features of the genotype, evolutionary process and phenotype are illustrated as an example.}
\label{fig.evocycle}
\end{centering}
\end{figure*}

The process of evolving plastic neural networks is depicted in Fig.\ \ref{fig.evocycle}. 

\section{Progress on evolving artificial plastic neural networks}
\label{sec:EAPB}

\begin{figure*}[t]
\centering
\includegraphics[width=0.9\textwidth]{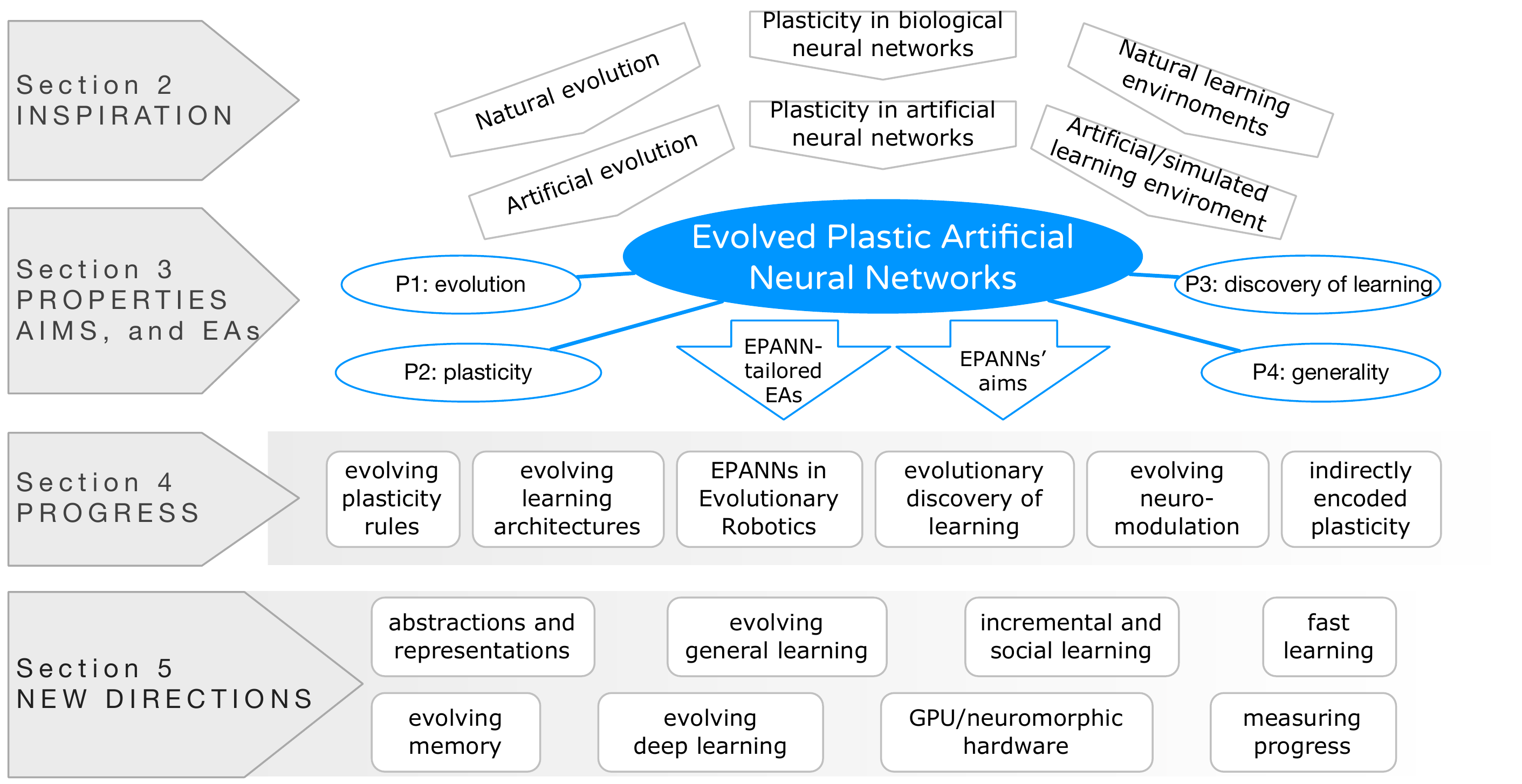}
\caption{Organization of the field of EPANNs, reflected in the structure of this paper.}
\label{fig.B2Lstructure}
\end{figure*}

This section reviews studies that have evolved plastic neural networks (EPANNs). The survey is divided into six sections that mirror the analysis of the field up to this point: the evolution of plasticity rules, the evolution of neural architectures, EPANNs in evolutionary robotics, the evolutionary discovery of learning, the evolution of neuromodulation, and the evolution of indirectly encoded plasticity. Accordingly, Figure \ref{fig.B2Lstructure} provides our perspective on the organization of the field, reflected in the structure of this paper.

\subsection{Evolving plasticity rules}
\label{sec:evo_rule}

Early EPANN experiments evolved the parameters of learning rules for fixed or hand-designed ANN architectures. Learning rules are functions that change the connection weight $w$ between two neurons, and are generally expressed as 
\begin{equation}
\Delta w = f(\mathbf{x},\mathbf{\theta})\enspace,
\label{eq.generalRule}
\end{equation} where $\mathbf{x}$ is a vector of neural signals and $\mathbf{\theta}$ is a vector of fixed parameters that can be searched by evolution. The incoming connection weights $\mathbf{w}$ to a neuron $i$ are used to determine the activation value
\begin{equation}
x_i = \sigma\big(\sum(w_{ji} \cdot x_j)\big)\enspace,
\label{eq.transferF}
\end{equation} where $x_j$ are the activation values of presynaptic neurons that connect to neuron $i$ with the weights $w_{ji}$, and $\sigma$ is a nonlinear function such as the sigmoid or the hyperbolic tangent. The vector $\mathbf{x}$ may provide local signals such as pre and postsynaptic activities, the value of the weight $w$, and modulatory or error signals. 

\cite{bengio1990learning,bengioBengioCloutierGecsei1992} proposed the optimization of the parameters $\theta$ of generic learning rules with gradient descent, simulated annealing, and evolutionary search for problems such as conditioning, boolean function mapping, and classification. Those studies are also among the first to include a modulatory term in the learning rules. The optimization was shown to improve the performance in those different tasks with respect to manual parameter settings. \citet{chalmers1990evolution} evolved a learning rule that applied to every connection and had a teaching signal. He found that, in 20\% of the evolutionary runs, the algorithm rediscovered, through evolution, the well-known delta rule, or Widrow-Hoff rule \citep{widrow1960adaptive}, used in backpropagation, thereby demonstrating the validity of evolution as an autonomous tool to discover learning. \citet{fontanari1991evolving} used the same approach of \citet{chalmers1990evolution} but constrained the weights to binary values. Also in this case, evolution autonomously rediscovered a hand-designed rule, the directed drift rule by \cite{venkatesh1993directed}. They also observed that the performance on new tasks was better when the network evolved on a larger set of tasks, possibly encouraging the evolution of more general learning strategies.  

With backpropagation of errors \citep{widrow199030}, the input vector $\mathbf{x}$ of Eq.\ \ref{eq.generalRule} requires an error signal between each input/output pair. In contrast, rules that use only local signals have been
a more popular choice for EPANNs, though this is changing with the rise in effectiveness of deep learning using back-propagation and related methods. In the simplest form, the product of presynaptic ($x_j$) and postsynaptic ($x_i$) activities, and a learning rate $\eta$ 
\begin{equation}
\Delta w = \eta \cdot x_j \cdot x_i
\label{eq.basicHebb}
\end{equation}
is known as Hebbian plasticity \citep{hebb1949,cooper2005}. More generally, any function as in Eq. \ref{eq.generalRule} that uses only local signals is considered a local plasticity rule for unsupervised learning. \citet{baxter1992} evolved a network that applied the basic Hebbian rule in Eq. \ref{eq.basicHebb} to a subset of weights (determined by evolution) to learn four functions of one variable. The network, called Local Binary Neural Net (LBNN), evolved to change its weights to one of two possible values ($\pm 1$), or have fixed weights. The experiment proved that learning can evolve when rules are optimized and applied to individual connections.

\citet{nolfi1993auto} evolved networks with ``auto-teaching'' inputs, which could then provide an error signal for the network to adjust weights during lifetime. The implication is that error signals do not always need to be hand-designed but can be discovered by evolution to fit a particular problem. A set of eight different local rules was used in \cite{rolls2000design} to investigate the evolution of rules in combination with the number of synaptic connections for each neuron, different neuron classes, and other network parameters. They found that evolution was effective in selecting specific rules from a large set to solve simple linear problems. In \cite{maniadakis2006modelling}, co-evolution was used to evolve agents (each being a network) that could use ten different types of Hebbian-like learning rules for simple navigation tasks: the authors reported that, despite the increase in the search space, using many different learning rules results in better performance but, understandably, a more difficult analysis of the evolved systems. \cite{meng2011modeling} evolved a gene regulatory network that in turn determined the learning parameters of the Bienenstock-Cooper-Munro (BCM) rule \citep{bienenstockCooperMunro1982}, showing promising performance in time series classification and other supervised tasks.

One general finding from these studies is that evolution operates well within large search spaces, particularly when a large set of evolvable rules is used.

\subsection{Evolving learning architectures}
\label{sec:evo_arch}

The interdependency of learning rules and neural architectures led to experiments in which evolution had more freedom on the network's design. The evolution of architectures in ANNs may involve searching an optimal number of hidden neurons, the number of layers in a network, particular topologies or modules, the type of connectivity, and other properties of the network's architecture. In EPANNs, evolving learning architectures implies more specifically to discover a combination of architectures and learning rules whose synergetic matching enables particular learning dynamics. As opposed to biological networks, EPANNs do not have the neurophysiological constraints, e.g., short neural
connections, sparsity, brain size limits, etc., that impose limitations on the natural evolution of biological networks. Thus, biologically implausible artificial systems may nevertheless be evolved in computer simulations \citep{bullinaria2007understanding,bullinaria2009importance}.

One seminal early study by \citet{happel1994design} proposed the evolutionary design of modular neural networks, called CALM \citep{murre1992learning}, in which modules could perform unsupervised learning, and the intermodule connectivity was shaped by Hebbian rules. The network learned categorization problems (simple patterns and hand written digits recognition), and showed that the use of evolution led to enhanced learning and better generalization capabilities in comparison to hand-designed networks. In \cite{arifovic2001using}, the authors used evolution to optimize the number of inputs and hidden nodes, and allowed connections in a feedforward neural network to be trained with backpropagation. \citet{abraham2004meta} proposed a method called Meta-Learning Evolutionary Artificial Neural Networks (MLEANN) in which evolution searches for initial weights, neural architectures and transfer functions for a range of supervised learning problems to be solved by evolved networks. The evolved networks were tested in time series prediction and compared with manually designed networks. The analysis showed that evolution consistently found networks with better performance than the hand-designed structures. \citet{khanMillerHalliday2008} proposed an evolutionary developmental system that created an architecture that adapted with learning: the network had a dynamic morphology in which neurons could be inserted or deleted, and synaptic connections formed and changed in response to stimuli. The networks were evolved with Cartesian genetic programming and appeared to improve their performance while playing checkers over the generations. \citet{downing2007gecco} looked at different computational models of neurogenesis to evolve learning architectures. The proposed evolutionary developmental system focused in particular on abstraction levels and principles such as Neural Darwinism \citep{edelman2000universe}. A combination of evolution of recurrent networks with a linear learner in the output was proposed in \citet{schmidhuber2007training}, showing that the evolved RNNs were more compact and resulted in better learning than randomly initialized echo state networks \citep{jaeger2004harnessing}. In \citet{KhanKM11,khanMillerHalliday2011,khan2014search}, the authors introduced a large number of bio-inspired mechanisms to evolve networks with rich learning dynamics. The idea was to use evolution to design a network that was capable of advanced plasticity such as dendrite branch and axon growth and shrinkage, neuron insertion and destruction, and many others. The system was tested on the Wumpus World \citep{russell2013AI}, a fairly simple problem with no learning required, but the purpose was to show that evolution can design working control networks even within a large search space.

In summary, learning mechanisms and neural architectures are strongly interdependent, but a large set of available dynamics seem to facilitate the evolution of learning.  Thus, EPANNs become more effective precisely when manual network design becomes less practical because of complexity and rich dynamics. 

\subsection{EPANNs in Evolutionary Robotics}
\label{sec:evo_rob}

Evolutionary robotics (ER) \citep{cliffHusbansHarvey1993,floreanoMondada1994,floreanoMondada1996,urzelaiFloreano2000,floreanoNolfi2004book} contributed strongly to the development of EPANNs, providing a testbed for applied controllers in robotics. Although ER had no specific assumptions on neural systems or plasticity \citep{smith02b}, robotics experiments suggested that neural control structures evolved with fixed weights perform less well than those evolved with plastic weights \citep{nolfi1996learning,floreanoUrzelai2001plastic}. In a conditional phototaxis robotic experiment\footnote{The fitness value was the time spent by a two-wheeled robot in one particular area of the area when a light was on, divided by the total experiment time.}, \cite{floreanoUrzelai2001morphogenesis} reported that networks evolved faster when synaptic plasticity and neural architectures were evolved simultaneously. In particular, plastic networks were shown to adapt better in the transition from simulation to real robots. The better simulation-to-hardware transition, and the increased adaptability in changing ER environments, appeared intuitive and supported by evidence \citep{nolfi1996learning}. However, the precise nature and magnitude of the changes from simulation to hardware is not always easy to quantify: those studies do not clearly outline the precise principles, e.g., better or adaptive feedback control, minimization principles, etc., that are discovered by evolution with plasticity to produce those advantages. In fact, the behavioral changes required to switch behaviors in simple ER experiments can also take place with non-plastic recurrent neural networks because evolution can discover recurrent units that act as switches. A study in 2003 observed similar performance in an associative learning task (food foraging) when comparing plastic and non-plastic recurrent networks \citep{stanleyMiikkulainen2003adaptive}. Recurrent networks with leaky integrators as neurons \citep{beerGallagher92,funahashi1993approximation,yamauchi94} were also observed to achieve similar performance to plastic networks \citep{blynelFloreano2002,blynelFloreano2003}. These early studies indicate that the evolution of learning with plastic networks was at that point still a proof-of-concept rather than a superior learning tool: aided by evolutionary search, networks with recurrent connections and fixed weights could create recurrent nodes, retain information and achieve similar learning performance to networks with plastic weights. 

Nevertheless, ER maintained a focus on plasticity as demonstrated, e.g., in The Cyber Rodent Project \citep{doyaUchibe2005} that investigated the evolution of learning by seeking to implement a number of features such as (1) evolution of neural controllers, (2) learning of foraging and mating behaviors, (3) evolution of learning architectures and meta-parameters, (4) simultaneous learning of multiple agents in a body, and (5) learning and evolution in a self-sustained colony. Plasticity in the form of modulated neural activation was used in \citet{husbandsSmithJakobiOshea1998} and \citet{smithHusbandsPhilippidesOShea2002} with a network that adapts its activation functions according to the diffusion of a simulated gas spreading to the substrate of the network. Although the robotic visual discrimination tasks did not involve learning, the plastic networks appeared to evolve faster than a network evolved with fixed activation functions. Similar conclusions were reached in \cite{dipaolo2003} and \cite{federici05b}. \cite{di2002spike,dipaolo2003} evolved networks with STDP for a wheeled robot to perform positive and negative phototaxis, depending on a conditioned stimulus, and observed that networks with fixed weights could learn but had inferior performance with respect to plastic networks. \cite{federici05b} evolved plastic networks with STDP and an indirect encoding, showing that plasticity helped performance even if learning was not required. Stability and evolvability of simple robotic controllers were investigated in \cite{hoinville2011flexible} who focused on EPANNs with homeostatic mechanisms.

Experiments in ER in the 1990s and early 2000s revealed the extent, complexity, and multitude of ideas behind the evolutionary design of learning neuro-robotics controllers. They generally indicate that plasticity helps evolution under a variety of conditions, even when learning is not required, thereby promoting further interest in more specific topics. 
Among those are the evolutionary discovery of learning, the evolution of neuromodulation, and the evolution of indirectly encoded plasticity, as described in the following.

\subsection{Evolutionary discovery of learning}
\label{sec:interaction_evo_learning}

When evolution is used to search for learning mechanisms, two main cases can be distinguished: (1) when learning is used to acquire constant facts about the agent or environment, and (2) when learning is used to acquire changeable facts. The first case, that of \emph{static} or \emph{stationary environments}, is known to be affected by the Baldwin effect \citep{baldwin1896} that suggests an acceleration of evolution when learning occurs during lifetime.  A number of studies showed that the Baldwin effect can be observed with computational simulations \citep{smith1986learning,hinton87,boers1995evolving,mayley1996landscapes,bullinaria2001exploring}. With static environments, learning causes a faster transfer of knowledge into the genotype, which can happen when facts are stationary (or constant) across generations. Eventually, a system in those conditions can perform well without learning because it can be born knowing to perform well. However, one limitation is that the genome might grow very large to hold large amount of information, and might, as a result, become less easy to evolve further. A second limitation is that such solutions might not perform well in non-stationary environments.

In the second case, that of \emph{variable} or \emph{non-stationary environments}, facts cannot be embedded in the genotype because those are changeable as, e.g., the location of food in a foraging problem. This case requires the evolution of learning for the performance to be maximized. For this reason, non-stationary reward-based environments, in which the behaviors to obtain rewards may change, are more typically used to study the evolution of learning in EPANNs.

\begin{figure}
\begin{centering}
\includegraphics[width=0.45\textwidth]{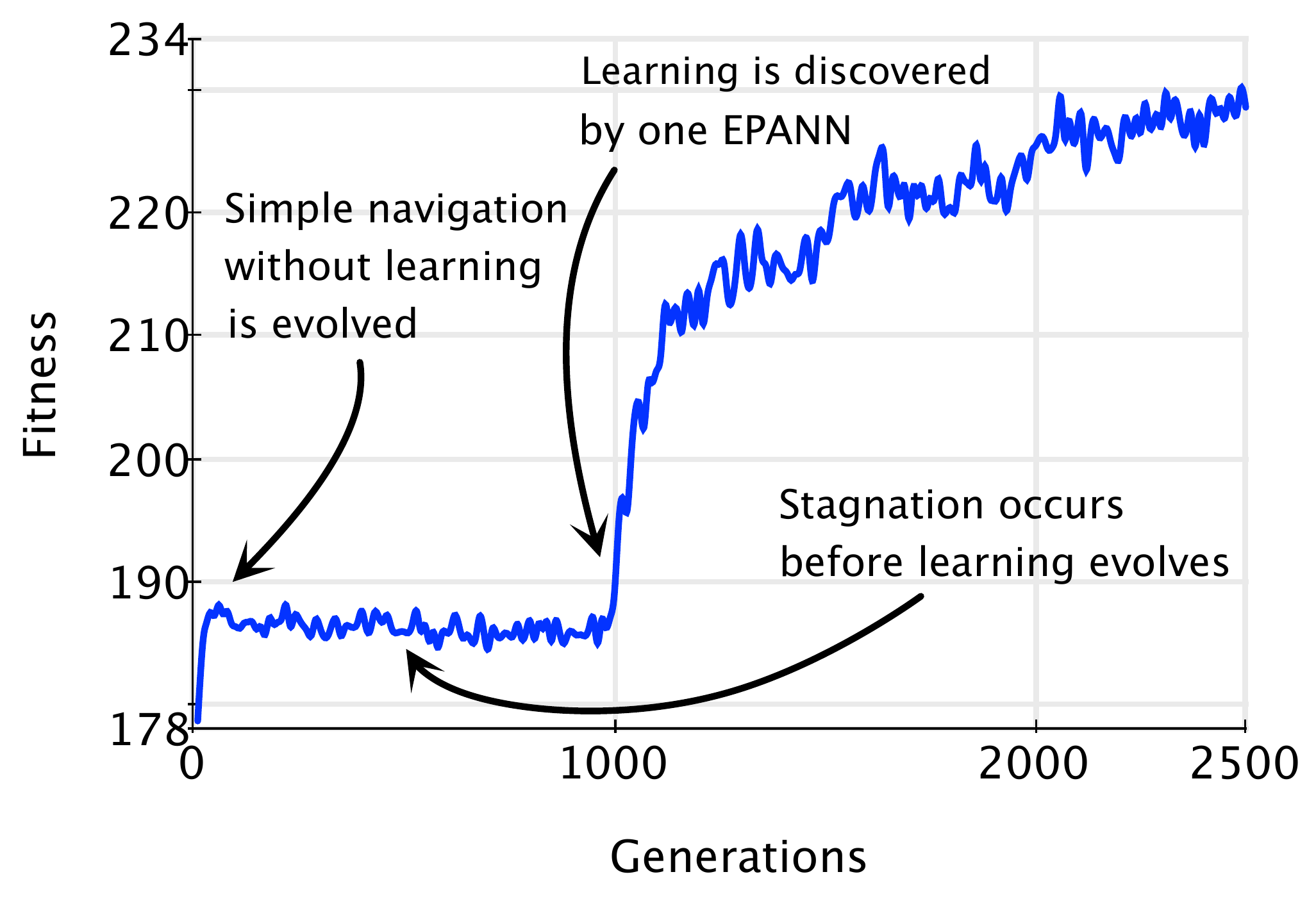}
\caption{{Discovery of a learning strategy during evolution in a non-stationary reward-based environment, i.e.\ where learning is required to maximise the fitness because of the variability of the environment. After approximately $1,000$ generations of $100$ individuals, a network evolves to use the reward signal to modify networks weights, and thus implement learning. The graphic is adapted from \citet{soltoggioDuerrMattiussiFloreano2007}. This experiment shows that evolution stagnates until learning is discovered. At that point, the evolutionary search hits on a gradient in the search space that improves on the learning as reflected by the fitness values.}}
\label{fig.fitnessJump}
\end{centering}
\end{figure}

EPANN experiments have been used to observe the advantages of combining learning and evolution, and the complex interaction dynamics that derives \citep{nolfi1999learning}. \cite{stone2007distributed} showed that distributed neural representations accelerate the evolution of adaptive behavior because learning part of a skill induced the automatic acquisition of other skill components. One study in a non-stationary environment (a foraging problem with variable rewards) \citep{soltoggioDuerrMattiussiFloreano2007} suggested that evolution discovers, before optimizing, learning in a process that is revealed by discrete fitness stepping stones. At first, non-learning solutions are present in the population. When evolution casually discovers a weak mechanism of learning, it is sufficient to create an evolutionary advantage, so the neural mechanism is subsequently optimized: Fig.\ \ref{fig.fitnessJump} 
shows a sudden jump in the fitness when one agent suddenly evolves a learning strategy: such jumps in fitness graphs are common in evolutionary experiments in which learning is discovered from scratch (\emph{Aim 1.1}), rather than optimized (\emph{Aim 1.2}), and were observed as early as in \cite{fontanari1991evolving}. When an environment changes over time, the frequency of those changes plays a role because it determines the time scales that are required from the learning agent. With time scales comparable to a lifetime, evolution may lead to phenotypic plasticity, which is the capacity for a genotype to express different phenotypes in response to different environmental conditions \citep{lalejinievolutionary2016}. The frequency of environmental changes was observed experimentally in plastic neural networks to affect the evolution of learning   \citep{ellefsen2014theEvolution}, revealing a complex relationship between environmental variability and evolved learning. One conclusion is that evolution copes with non-stationary environments by evolving the specific learning that better matches those changes. 

The use of reward to guide the discovery of neural learning through evolution was shown to be inherently deceptive in \citet{risi2010evolving} and \citet{lehman:gecco14}. In \citet{risi2009novelty,risi2010evolving}, EPANN-controlled simulated robots, evolved in a discrete T-Maze domain, revealed that the stepping stones towards discovering learning are often not rewarded by objective-based performance measures. Those stepping stones to learning receive a lower fitness score than more brittle solutions with no learning but effective behaviors. A solution to this problem  was devised in \citet{risi2010evolving,risi2009novelty}, in which  \emph{novelty search} \citep{lehman2008exploiting,lehman2011abandoning} was adopted as a substitute for performance in the fitness objective with the aim of finding novel behaviors. Novelty search  was observed to perform significantly better in the T-Maze domain. \citet{lehman:gecco14} later showed that novelty search can encourage the evolution of more adaptive behaviors across a variety of different variations of the T-Maze learning tasks. As a consequence, novelty search contributed to a philosophical change by questioning the centrality of objective-driven search in current evolutionary algorithms \citep{stanley2015greatness}. By rewarding novel behaviors, novelty search validates the importance of exploration or curiosity, previously proposed in \citet{schmidhuber1991curious,schmidhuber2006developmental}, also from an evolutionary viewpoint. With the aim of validating the same hypothesis, \citet{soltoggio2009novelty} devised a simple EPANN experiment in which exploration was more advantageous than exploitation in the absence of reward learning; to do this, the reward at a particular location depleted itself if continuously visited, so that changing location at random in a T-maze became beneficial. Evolution discovered exploratory behavior before discovering reward-learning, which in turn, and surprisingly, led to an earlier evolution of reward-based learning. Counterintuitively, this experiment suggests that a stepping stone to evolve reward-based learning is to encourage reward-independent exploration.

 The seminal work in \cite{bullinaria2003biological,bullinaria2007effect,bullinaria2009lifetime} proposes the more general hypothesis that learning requires the evolution of long periods of parental protection and late onset of maturity. Similarly, \citet{ellefsen2013evolved,ellefsen2013different} investigates sensitive and critical periods of learning in evolved neural networks. This fascinating hypothesis has wider implications for experiments with EPANNs, and more generally for machine learning and AI. It is therefore foreseeable that future EPANNs will have a protected childhood during which parental guidance may be provided \citep{clutton1991evolution,klug2010life,eskridge2012nurturing}.

\subsection{Evolving neuromodulation}
\label{sec:NM}

Growing neuroscientific evidence on the role of neuromodulation (previously outlined in Section \ref{sec:plasticity_in_biological}) inspired the design of experiments with neuromodulatory signals to evolve control behavior and learning strategies (Section \ref{sec:plasticity_in_artificial}). One particular case is when neuromodulation gates plasticity. Eq.\ \ref{eq.generalRule} can be rewritten as as 
\begin{equation}
\Delta w = m\cdot f(\mathbf{x},{\bm{\theta}})\enspace,
\label{eq.modGen}
\end{equation}
to emphasize the role of $m$, a modulatory signal used as a multiplicative factor that can enhance or reduce plasticity \citep{abbott1990}. A network may produce many independent modulatory signals $\mathbf{m}$ targeting different neurons or areas of the network. Thus, modulation can vary in space and time. Modulation may also affect other aspects of the network dynamics, e.g., modulating activations rather than plasticity \citep{krichmar2008neuromodulatory}. Graphically, modulation can be represented as a different type of signal affecting various properties of the synaptic connections of an afferent neuron $i$ \begin{figure}
\begin{centering}
\includegraphics[width=0.50\textwidth]{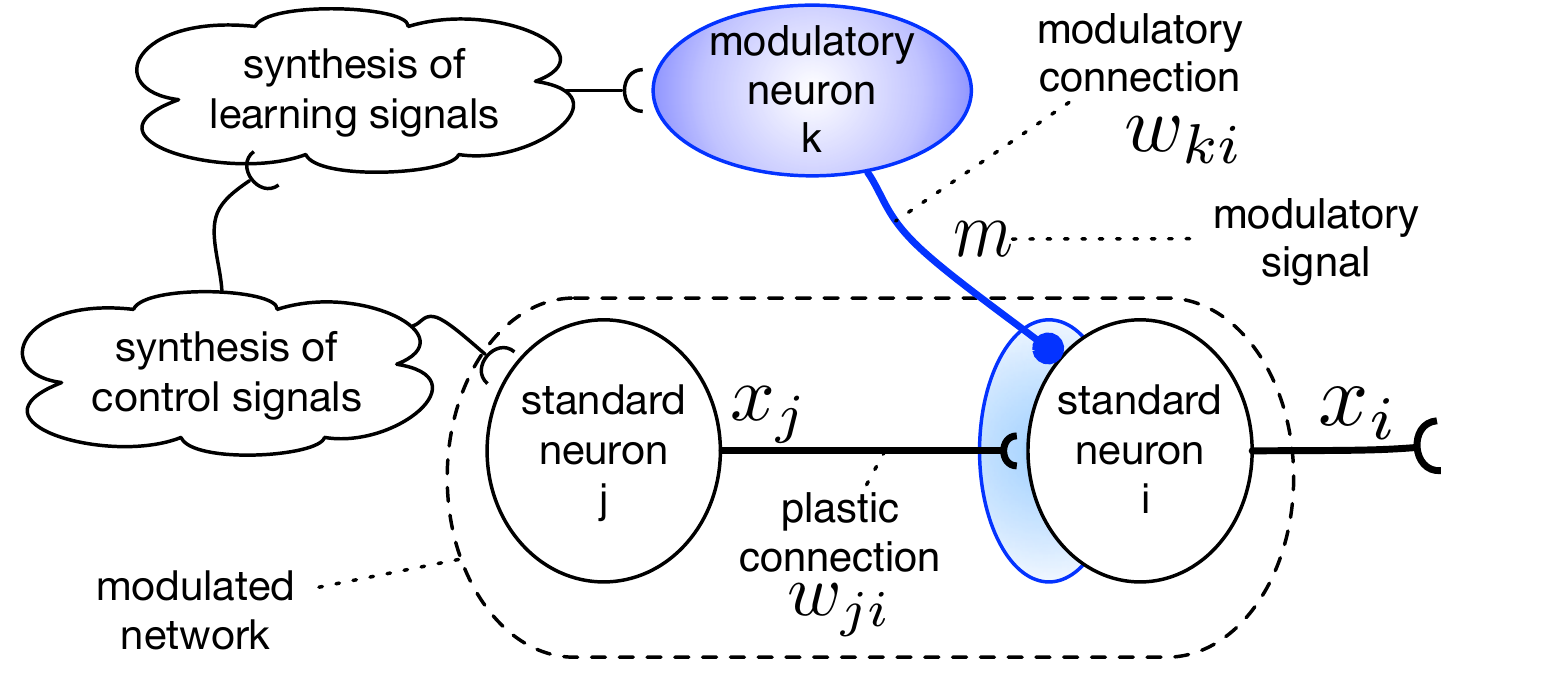}
\caption{A modulatory neuron gates plasticity of the synapses that connect to the postsynaptic neuron. The learning is local, but a learning signal can be created by one part of the network and used to regulate learning elsewhere.}
\label{fig.mod2}
\end{centering}
\end{figure} (Fig.\ \ref{fig.mod2}).

Evolutionary search was used to find the parameters of a neuromodulated Hebbian learning rule in a reward-based armed-bandit problem in \citet{nivJoelMeilijsonRuppin2002}. The same problem was used later in \citet{soltoggioDuerrMattiussiFloreano2007} to evolve arbitrary learning architectures with a bio-inspired gene representation method called Analog Genetic Encoding (AGE) \citep{mattiussi2007analog}. In that study, evolution was used to search both  modulatory topologies and parameters of a particular form of Eq.\ \ref{eq.modGen}:
\begin{equation}
\Delta w = m\cdot (Ax_i x_j + Bx_j + Cx_i +D)\enspace,
\label{eq.modABCD}
\end{equation}
where the parameters $A$ to $D$ determined the influence of four factors in the rule: a multiplicative Hebbian term $A$, a presynaptic term $B$, a postsynaptic term $C$, and pure modulatory, or heterosynaptic, term $D$. Such a rule is not dissimilar from those presented in previous studies (see Section 3.2). However, when used in combination with modulation and a search for network topologies, evolution seems to be particularly effective at solving reward-based problems. \citet{kondo2007} proposed an evolutionary design and behavior analysis of neuromodulatory neural networks for mobile robot control, validating the potential of the method.

\citet{soltoggioBullinariaMattiussiDuerrFloreano2008} tested the question of whether modulatory dynamics held an evolutionary advantage in T-maze environments with changing reward locations\footnote{In reward-based T-Maze environments, it is often assumed that the fitness function is the sum or all rewards collected during a lifetime.}. In their algorithm, modulatory neurons were freely inserted or deleted by random mutations, effectively allowing the evolutionary selection mechanism to autonomously pick those networks with advantageous computational components (\emph{Aim 2}). After evolution, the best performing networks had modulatory neurons regulating learning, and evolved faster than a control evolutionary experiment that could not employ modulatory neurons. Modulatory neurons were maintained in the networks in a second phase of the experiment when genetic operators allowed for the deletion of such neurons but not for their insertion, thus demonstrating their essential function in maintaining learning in that particular experiment. In another study,  \citet{soltoggio2008epi} suggested that evolved modulatory topologies may be essential to separate the learning circuity from the input-output controller, and shortening the input-output pathways which sped up decision processes.  \citet{soltoggio2008HIS} showed that the learning dynamics are affected by tight coupling between rules and architectures in a search space with many equivalent but different control structures. Fig.\ \ref{fig.mod2} also suggests that modulatory networks require evolution to find two essential topological structures: what signals or combination of signals trigger modulation, and what neurons are to be targeted by modulatory signals. In other words, a balance between fixed and plastic architectures, or selective plasticity \citep{DARPA-BAA-HR001117S0016}, is an intrinsically emergent property of evolved modulated networks. 

A number of further studies on the evolution of neuromodulatory dynamics confirmed the evolutionary advantages in learning scenarios \citep{soltoggioPhDThesis2008}. \citet{silva2012adaptation} used simulations of 2-wheel robots performing a dynamic concurrent foraging task, in which scattered food items periodically changed their nutritive value or became poisonous, similarly to the setup in \citet{soltoggioStanley2012}. The results showed that when neuromodulation was enabled, learning evolved faster than when neuromodulation was not enabled, also with multi-robot distributed systems \citep{silva2012continuous}. \citet{nogueira2013evolving,nogueira2016emergent} also reported evolutionary advantages in foraging behavior of an autonomous virtual robot when equipped with neuromodulated plasticity. \citet{harrington2013robot} demonstrated how evolved neuromodulation applied to a gene regulatory network consistently generalized better than agents trained with fixed parameter settings. Interestingly, \citet{arnold2013selection} showed that neuromodulatory architectures provided an evolutionary advantage also in reinforcement-free environments, validating the hypothesis that plastic modulated networks have higher evolvability in a large variety of tasks. The evolution of social representations in neural networks was shown to be facilitated by neuromodulatory dynamics in \citet{arnold2013evolution}. An artificial life simulation environment called Polyworld \citep{yoderYaeger2014} helped to assess the advantage of neuromodulated plasticity in various scenarios. The authors found that neuromodulation may be able to enhance or diminish foraging performance in a competitive, dynamic environment. 

Neuromodulation was evolved in \citet{ellefsen2015neural} in combination with modularity to address the problem of catastrophic forgetting. In \citet{GustafssonMsc2016}, networks evolved with AGE \citep{mattiussi2007analog} for video game playing were shown to perform better with the addition of neuromodulation. \citet{norouzzadehClune} showed that neuromodulation produced forward models that could adapt to changes significantly better than the controls. They verified that evolution exploited variable learning rates to perform adaptation when needed. In \cite{DBLP:journals/corr/VelezC17}, diffusion-based modulation, i.e., targeting entire parts of the network, evolved to produce task-specific localized learning and functional modularity, thus reducing the problem of catastrophic forgetting.

The evidence in these studies suggests that neuromodulation is a key ingredient to facilitate the evolution of learning in EPANNs. They also indirectly suggest that neural systems with more than one type of signal, e.g., activation and other modulatory signals, might be beneficial in the neuroevolution of learning.

\subsection{Evolving indirectly encoded plasticity}

\begin{figure*}
\centering
\subfloat[][HyperNEAT]
{
\includegraphics[width=0.5\textwidth]{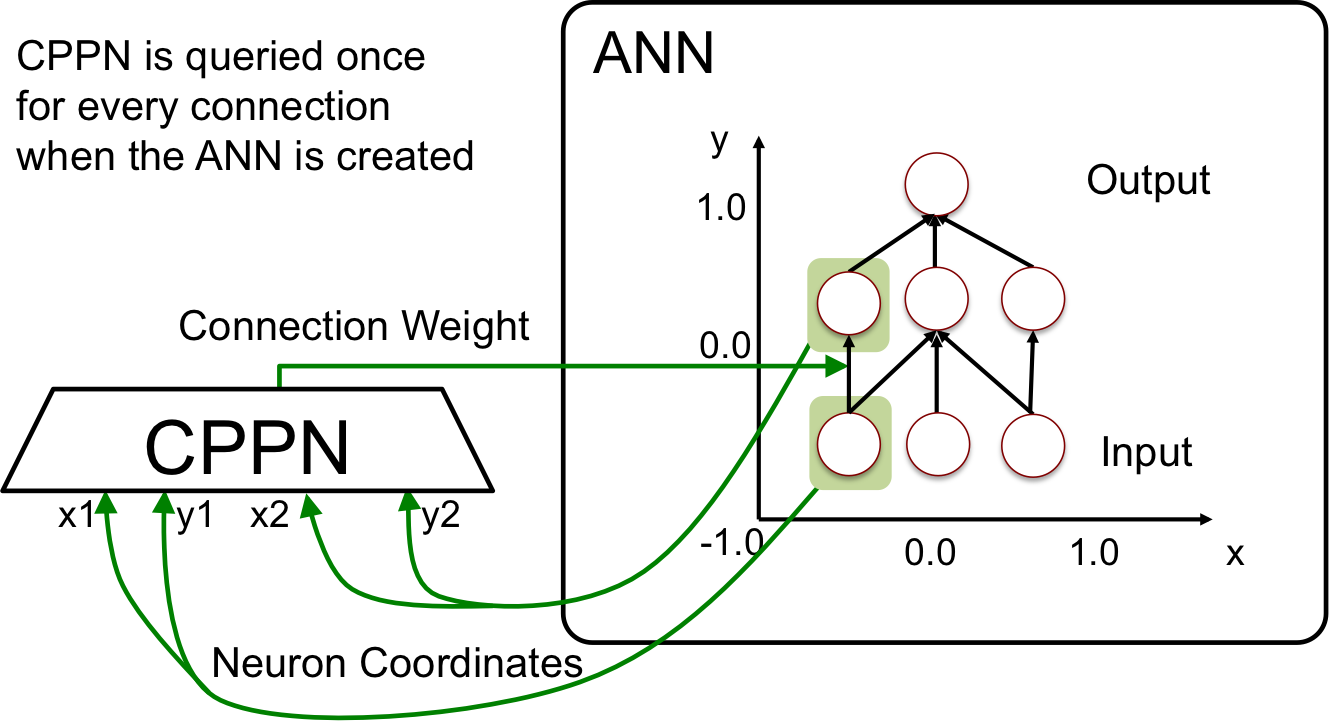}
}\\
\subfloat[][Adaptive HyperNEAT]
{
\includegraphics[width=0.85\textwidth]{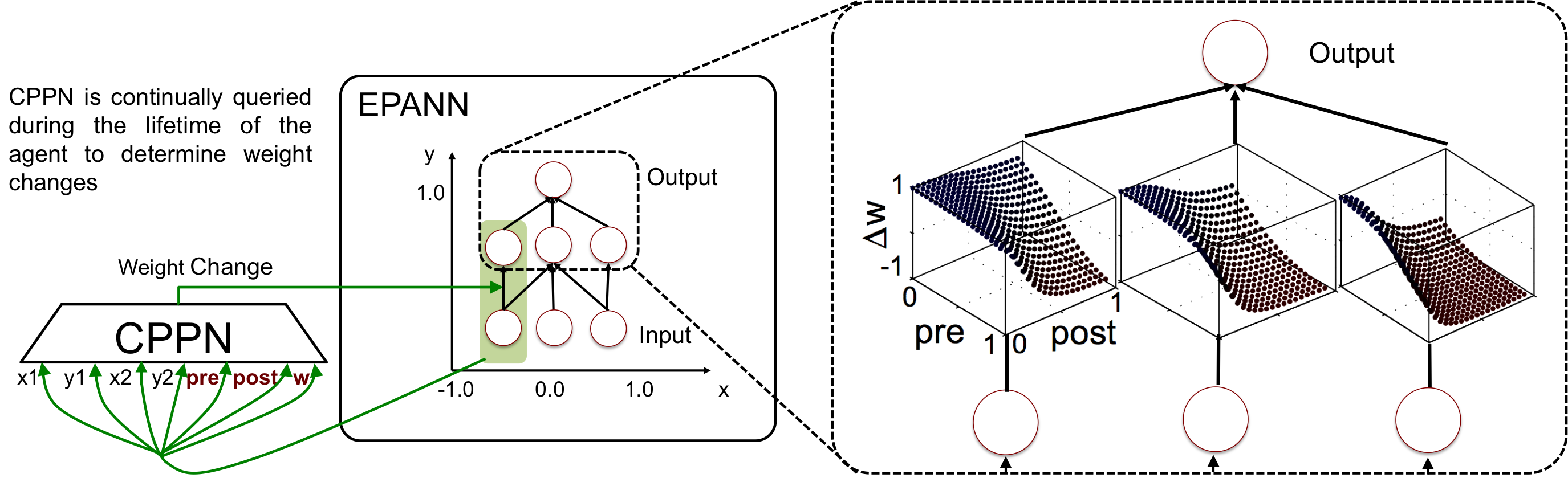}
}
\caption{{Example of an indirect mapping of plasticity rules from a compact genotype to a larger phenotype.  (a) ANN nodes in HyperNEAT are situated in physical space by assigning them specific coordinates. The connections between nodes are determined by an evolved Compositional Patterns Producing Network (CPPN;~\cite{stanley2007compositional}), which takes as inputs the coordinates of two ANN neurons and returns the weight between them. In the normal HyperNEAT approach (a), the CPPN is queried once for all potential ANN connections when the agent is born.  On the other hand, in adaptive HyperNEAT (b), the CPPN is continually queried during the lifetime of the agent to determine individual connection weight changes based on the location of neurons and additionally the activity of the presynaptic and postsynaptic neuron, and current connection weight. Adaptive HyperNEAT is able to indirectly encode a pattern of nonlinear learning rules for each connection in the ANN (right).}}
 \label{fig:hyperneat}
\end{figure*}

An indirect genotype to phenotype mapping means that evolution operates on a compact genotypical representation (analogous to the DNA) that is then mapped into a fully fledged network (analogous to a biological brain). Learning rules may undergo a similar indirect mapping, so that compact instructions in the genome expand to fully fledged plasticity rules in the phenotype. One early study \citep{gruau1993adding} encoded plasticity and development with a grammar tree, and compared different learning rules on a simple static task (parity and symmetry), demonstrating that learning provided an evolutionary advantage in a static scenario. In non-static contexts, and using a T-Maze domain as learning task, \cite{risi2010indirectly} showed that HyperNEAT, which usually implements a compact encoding of  weight patterns for large-scale ANNs (Fig.~\ref{fig:hyperneat}a), can also encode patterns of local learning rules. The approach, called \emph{adaptive HyperNEAT}, can encode arbitrary learning rules for each connection in an evolving ANN based on a function of the ANN's geometry (Fig.~\ref{fig:hyperneat}b). Further flexibility was added in  \citet{risi2012unified} to simultaneously encode the density and placement of nodes in substrate space. The approach, called  \emph{adaptive evolvable-substrate HyperNEAT}, makes it possible to indirectly encode plastic ANNs with thousands of connections that exhibit regularities and repeating motifs. Adaptive ES-HyperNEAT allows each individual synaptic connection, rather than neuron, to be standard or  modulatory, thus introducing further design flexibility.  \citet{risiStanley2014} showed how adaptive HyperNEAT can be seeded to produce a specific lateral connectivity pattern, thereby allowing the weights to self-organize to form a topographic map of the input space. The study shows that evolution can be seeded with specific plasticity mechanisms that can facilitate the evolution of specific types of learning. 

The effect of indirectly encoded plasticity on the learning and on the evolutionary process was investigated by \citet{tonelli2011using,tonelli2013relationships}. Using an operant conditioning task, i.e., learning by reward, the authors showed that indirect encodings that produced more regular neural structures also improved the general EPANN learning abilities when compared to direct encodings. In an approach similar to adaptive HyperNEAT, \citet{orchard2016evolution} encoded the learning rule itself as an evolving network. They named the approach \emph{neural weights and bias update} (NWB), and observed that increasing the search space of the possible plasticity rules created more general solutions than those based on only Hebbian learning.


\section{Future directions}
\label{sec:OC}

The progress of EPANNs reviewed so far is based on rapidly developing theories and technologies. In particular, new advances in AI, machine learning, neural networks and increased computational resources are currently creating a new fertile research landscape, and are setting the groundwork for new directions for EPANNs. This section presents promising research themes that have the potential to extend and radically change the field of EPANNs and AI as a whole.

\subsection{Levels of abstraction and representations}

Choosing the right level of abstraction and the right representation \citep{bengio2013representation} are  themes at the heart of many problems in AI. In ANNs, low levels of abstraction are more computationally expensive, but might be richer in dynamics. High levels are faster to simulate, but require an intuition of the essential dynamics that are necessary in the model. Research in EPANNs is well placed to address the problem of levels of abstraction because it can reveal evolutionary advantages for different components, structures and representations.

Similarly to abstractions, representations play a critical role. Compositional Patterns Producing Networks (CPPNs) \citep{stanley2007compositional}, and also the previous work of \citet{sims1991artificial}, demonstrated that structured phenotypes can be generated through a function without going through the dynamic developmental process typical of multicellular organisms. Relatedly, \citet{hornby2002creating}  showed that the different phenotypical representations led to considerably different results in the evolution of regular structures with patterns and repetitions. \citet{miller2014neuro} discussed explicitly the effect of abstraction levels for evolved developmental learning networks, in particular in relation to two approaches that model development at the neuron level or at the network level. 

Finding appropriate abstractions and representations, just as it was fundamental in the advances in deep learning to represent input spaces and hierarchical features \citep{bengio2013representation,oquab2014learning}, can also extend to representations of internal models, learning mechanisms, and genetic encodings, affecting the algorithms' capabilities of evolving learning abilities. 

\subsection{Evolving general learning}

One challenge in the evolution of learning is that evolved learning may simply result in a switch among a finite set of evolved behaviors, e.g., turning left or right in a T-Maze in a finite sequence, which is all that evolving solutions encounter during their lifetime. A challenge for EPANNs is to acquire general learning abilities in which the network is capable of learning problems not encountered during evolution. \citet{mouret2014artificial} propose the distinction between the evolution of behavioral switches and the evolution of synaptic general learning abilities, and suggest conditions that favor these types of learning. General learning can be intuitively understood as the capability to learn any association among input, internal, and output patterns, both in the spatial and temporal dimensions, regardless of the complexity of the problem. Such an objective clearly poses practical and philosophical challenges. Although humans are considered better at general learning than machines, human learning skills are also specific and not unlimited \citep{ormrod2004human}.  Nevertheless, moving from behavior switches to more general learning is a desirable feature for EPANNs. Encouraging the emergence of general learners may likely involve (1) an increased computational cost for testing in rich environments that include a large variety of uncertain and stochastic scenarios with problems of various complexity, and (2) an increased search space to explore the evolution of complex strategies and avoid deception. 

\subsection{Incremental and social learning} 
\label{sec:continual}
An important open challenge for machine learning in general is the creation of neural systems that can continuously integrate new knowledge and skills without forgetting what they previously learned \citep{parisi2018continual}, thus solving the stability-plasticity dilemma. A promising approach is progressive neural networks \citep{rusu2016progressive}, in which a new network is created for each new task, and lateral connections between networks allow the system to leverage previously learned features. In the presence of time delays among stimuli, actions and rewards, a rule called hypothesis testing plasticity (HTP) \citep{soltoggioHTP2014} implements fast and slow decay to consolidate weights and suggests neural dynamics to avoid catastrophic forgetting. A method to find the best shared weights across multiple tasks, called elastic weight consolidation (EWC) was proposed in \citet{Kirkpatrick14032017}. Plasticity rules that implement weight consolidation, given their promise to prevent catastrophic forgetting, are likely to become standard components in EPANNs.

Encouraging modularity  \citep{ellefsen2015neural,durr2010genetic} or augmenting evolving networks with a dedicated external memory component \citep{lueders2016NIPS} have been proposed recently. An evolutionary advantage is likely to  emerge for networks that can elaborate on previously learned sub-skills during their lifetime to learn more complex tasks. 

One interesting case in which incremental learning may play a role is social learning \citep{best1999culture}. EPANNs may learn both from the environment and from other individuals, from scratch or incrementally \citep{offerman1998learning}. In an early study, \citet{mcquesten1997culling} showed that neuroevolution can
benefit from parent networks teaching their offspring through backpropagation. When social, learning may involve imitation, language or communication, or other social behaviors. \citet{bullinaria2017imitative} proposes an EPANN framework to simulate the evolution of culture and social learning. It is reasonable to assume that future AI learning systems, whether based on EPANNs or not, will acquire knowledge through different modalities. These will involve direct experience with the environment, but also social interaction, and possibly complex incremental learning phases.

\subsection{Fast learning} 

Animal learning does not always require a myriad of trials. Humans can very quickly generalize from only a few given examples, possibly leveraging previous experiences and a long learning process during infancy. This type of learning, advocated in AI and robotics systems \citep{thrun1995lifelong}, is currently still missing in EPANNs. Inspiration for new approaches could come from complementary learning systems \citep{mcclelland1995there,kumaran2016learning} that humans seem to possess, which include fast and slow learning components. Additionally, approaches such as probabilistic program induction seem to be able to learn concepts in one-shot at a human-level in some tasks \citep{lake2015human}.
Fast learning is likely to derive not just from trial-and-error, but also from mental models that can be applied to diverse problems, similarly to transfer learning \citep{thrun1995lifelong,thrun1996discovering,pan2010survey}. Reusable mental models, once learned, will allow agents to make predictions and plan in new and uncertain scenarios with similarities to previously learned ones. If EPANNs can discover neural structures or learning rules that allow for generalization, an evolutionary advantage of such a discovery will lead to its full emergence and further optimization of such a property. 

A rather different approach to accelerate learning was proposed in \cite{fernando2008copying,de2015neuronal} and called Evolutionary Neurodynamics. According to this theory, replication and selection might happen in a neural system as it learns, thus mimicking an evolutionary dynamics at the much faster time scale of a lifetime. We refer to \cite{fernando2012selectionist,de2015neuronal} for an overview of the field. The appeal of this method is that evolutionary search can be accelerated by implementing its dynamics at both the evolution's and life's time scales.

\subsection{Evolving memory} 

\label{sec:storing}
The consequence of learning is memory, both explicit and implicit \citep{anderson2013architecture}, and its consolidation \citep{dudai2012restless}. For a review of computational models of memory see \cite{fusi2017computational}. EPANNs may reach solutions in which memory evolved in different fashions, e.g., preserved as self-sustained neural activity, encoded by connection weights modified by plasticity rules, stored with an external memory (e.g.\ Neural Turing Machine), or a combination of these approaches.  Recurrent neural architectures based on \emph{long short-term memory} (LSTM) allow very complex tasks to be solved through gradient descent training \citep{greff2015lstm,hochreiter1997long} and have recently shown promise when combined with evolution \citep{Rawal2016lstm}. Neuromodulation and weight consolidation could also be used to target areas of the network where information is stored.

\citet{graves2014neural} introduced the \emph{Neural Turing Machine} (NTM), networks augmented with an external memory that allows long-term memory storage. NTMs have shown promise when trained through evolution \citep{Greve2016ENTM, lueders2016NIPS, lueders2017evostar} or gradient descent \citep{graves2014neural, graves2016hybrid}. The \emph{Evolvable Neural Turing Machine} (ENTM) showed good performance in solving the continuous version of the double T-Maze navigation task \citep{Greve2016ENTM}, and avoided catastrophic forgetting in a continual learning domain \citep{lueders2016NIPS, lueders2017evostar} because memory and control are separated by design. Research in this area will reveal which computational systems are more evolvable and how memories will self organize and form in EPANNs. 

\subsection{EPANNs and deep learning}

Deep learning has shown remarkable results in a variety of different fields \citep{krizhevsky2012imagenet,schmidhuber2015deep,lecun2015deep}. However, the model structures of these networks are mostly hand-designed, include a large number of parameters, and require extensive experiments to discover optimal configurations. With increased computational resources, it is now possible to search design aspects with evolution, and set up EPANN experiments with the aim of optimizing learning (\emph{Aim 1.2}).

\citet{koutnik2014evolving} used evolution to design a controller that combined evolved  recurrent neural networks, for the control part, and a deep max-pooling convolutional neural network to reduce the input dimensionality. The study does not use evolution on the deep preprocessing networks itself, but demonstrates nevertheless the evolutionary design of a deep neural controller. \citet{young2015optimizing} used an evolutionary algorithm to optimize two parameters of a deep network: the  size (range [1,8]) and the number (range [16,126]) of the filters in a convolutional neural network, showing that the optimized parameters could vary considerably from the standard best-practice values. An established evolutionary computation technique, the Covariance Matrix Adaptation Evolution Strategy (CMA-ES) \citep{hansen2001completely}, was used in \citet{loshchilov2016cma} to optimize the parameters of a deep network to learn to classify the MNIST dataset. The authors reported performance close to the state-of-the-art using 30 GPU devices.  

\citet{real2017} and \cite{miikkulainen2017evolving} showed that evolutionary search can be used to determine the topology, hyperparameters and building blocks of deep networks trained through gradient descent. The performance were shown to rival those of hand-designed architectures in the CIFAR-10 classification task and a language modeling task \citep{miikkulainen2017evolving}, while \citet{real2017} also tested the method on the larger CIFAR-100 dataset. \cite{desell2017large} proposes a method called evolutionary exploration of augmenting convolutional topologies, inspired by NEAT \citep{stanley2002evolving}, which evolves progressively more complex unstructured convolutional neural networks using genetically specified feature maps and filters. This approach is also able to co-evolve neural network training hyperparameters. Results were obtained using 5,500 volunteered computers at the Citizen Science Grid who were able to evolve competitive results on the MNIST dataset in under 12,500 trained networks in a period of approximately ten days. \cite{liu2017hierarchical} used evolution to search for hierarchical architecture representations showing competitive performance on the CIFAR-10 and Imagenet databases. The Evolutionary DEep Networks (EDEN) framework \citep{dufourq2017eden} aims to generalize deep network optimization to a variety of problems and is interfaced with TensorFlow \citep{abadi2016tensorflow}. A number of similar software frameworks are currently being developed. 

\citet{fernando2017pathnet} used evolution to determine a subset of pathways through a network that are trained through backpropagation, allowing the same network to learn a variety of different tasks. \citet{fernando2016convolution} were also able to rediscover convolutional networks by means of evolution of Differentiable Pattern Producing Networks \citep{stanley2007compositional}. 

So far, EPANN experiments in deep learning have focused primarily on the optimization of learning (\emph{Aim 1.2}) in supervised classification tasks, e.g.\ optimizing final classification accuracy. In the future, evolutionary search may be used with deep networks to evolve learning from scratch, recover performance, or combining different learning rules and dynamics in an innovative and counter-intuitive fashion (\emph{Aims 1.1, 1.3 or 2} respectively). 

\subsection{GPU implementations and neuromorphic hardware} 

The progress of EPANNs will crucially depend on implementations that take advantage of the increased computational power of parallel computation with GPUs and neuromorphic hardware \citep{jo2010nanoscale,monroe2014neuromorphic}.  Deep learning greatly benefited from GPU-accelerated machine learning but also standardized tools (e.g.\ Torch, Tensorflow, Theano, etc.) that made it easy for anybody to download, experiment, and extend promising deep learning models.

{EPANNs have shown promise with hardware implementations.} \citet{howard2011towards,howard2012evolution,howard2014evolving} devised experiments to evolve plastic spiking networks implemented as memristors for  simulated robotic navigation tasks. Memristive plasticity was observed consistently to enable higher performance than constant-weighted connections in both static and dynamic reward scenarios. \cite{carlson2014efficient} used GPU implementations to evolve plastic spiking neural networks with an evolution strategy, which resulted in an efficient and automated parameter tuning framework. 

In the context of newly emerging technologies, it is worth noting that, just as GPUs were not developed initially for deep learning, so novel neural computation tools and hardware systems, not developed for EPANNs, can now be exploited to enable more advanced EPANN setups.

\subsection{Measuring progress} 
\label{sec.metrics}

{The number of platforms and environments for testing the capabilities of intelligent systems is constantly growing, e.g., the Atari or General Video Game Playing Benchmark \citep{gvgai.net}, the Project Malmo \citep{malmo}, or the OpenAI Universe \citep{openAI.com}.  Because EPANNs are often evolved in reward-based, survival, or novelty-oriented environments to discover new, unknown, or creative learning strategies or behaviors, measuring progress is not straightforward.} Desired behaviors or errors are not always defined. Moreover, the goal for EPANNs is often not to be good at solving one particular task, but rather to test the capability to evolve the learning required for a range of problems, to generalize to new problems, or to recover performance after a change in the environment. Therefore, EPANNs will require the community to devise and accept new metrics based on one or more objectives such as the following: \begin{itemize} 
\item the time (in the evolutionary scale) to evolve the learning mechanisms in one or more scenarios;
\item the time (in the lifetime scale) for learning in one or more scenarios;
\item the number of different tasks that an EPANN evolves to solve;
\item a measure of the variety of skills acquired by one EPANN;
\item the complexity of the tasks and/or datasets, e.g., variations in distributions, stochasticity, etc.;
\item the robustness and generalization capabilities of the learner;
\item the recovery time in front of high-level variations or changes, e.g., data distribution, type of problem, stochasticity levels, etc.;
\item computational resources used, e.g., number of lifetime evaluations, length of a lifetime;
\item size, complexity, and computational requirements of the solution once deployed;
\item novelty or richness of the behavior repertoire from multiple solutions, e.g., the variety of different EPANNs and their strategies that were designed during evolution.
\end{itemize}

Few of those metrics are currently used to benchmark machine learning algorithms. Research in EPANNs will foster the adoption of such criteria as wider performance metrics for assessing lifelong learning capabilities \citep{thrun2012learning,DARPA-BAA-HR001117S0016} of evolved plastic networks.

\vspace{-2pt}
\section{Conclusion}

The broad inspiration and aspirations of evolved artificial plastic neural networks (EPANNs) strongly motivate this field, drawing from large, diverse, and interdisciplinary areas. In particular, the aspirations reveal ambitious and long-term research objectives related to the discovery of neural learning, with important implications for artificial intelligence and biology.

EPANNs saw considerable progress in the last two decades, primarily pointing to the potential of the autonomous evolution and discovery of neural learning. We now have: (i) advanced evolutionary algorithms to promote the evolution of learning, (ii) a better understanding of the interaction dynamics between evolution and learning, (iii) assessed advantages of multi-signal networks such as modulatory networks, and (iv) explored evolutionary representations of learning mechanisms.

Recent scientific and technical progress has set the foundation for a potential step change in EPANNs. Concurrently with the increase of computational power and a resurgence of neural computation, the need for more flexible algorithms and the opportunity to explore new design principles could make EPANNs the next AI tool capable of discovering new principles and systems for general adaptation and intelligent systems. 

\vspace{-6pt}
\section*{Acknowledgements}

We thank John Bullinaria, Kris Carlson, Jeff Clune, Travis Desell, Keith Downing, Dean Hougen, Joel Lehman, Jeff Krichmar, Jay McClelland, Robert Merrison-Hort, Julian Miller, Jean-Baptiste Mouret, James Stone, Eors Szathmary, and Joanna Turner for insightful discussions and comments on earlier versions of this paper. 

\Urlmuskip=0mu plus 1mu\relax
\setlength{\bibsep}{1pt plus 0.3ex}
\balance

\end{document}